\documentclass[11pt,a4paper]{article}
\usepackage[hyperref]{acl2021}
\usepackage{times}
\usepackage{latexsym}

\usepackage{graphicx}
\usepackage{amsmath}
\usepackage{multirow}
\usepackage{amsthm}
\usepackage{amsfonts}
\usepackage{bm}
\usepackage{booktabs}
\usepackage{algorithm}
\usepackage{algorithmic}
\usepackage{verbatim}

\usepackage{microtype}

\aclfinalcopy 

\title{Search from History and Reason for Future: Two-stage Reasoning on Temporal Knowledge Graphs}

 \author{Zixuan Li\textsuperscript{1,2}, Xiaolong Jin\textsuperscript{1,2}, Saiping Guan\textsuperscript{1,2}, Wei Li\textsuperscript{3}, Jiafeng Guo\textsuperscript{1,2}, \\
 \textbf{Yuanzhuo Wang\textsuperscript{1,2} and Xueqi Cheng\textsuperscript{1,2}} \\
  \textsuperscript{1}School of Computer Science and Technology, University of Chinese Academy of Sciences; \\
  \textsuperscript{2}CAS Key Laboratory of Network Data Science and Technology, Institute of\\ Computing Technology, Chinese Academy of Sciences;\\
  \textsuperscript{3}Baidu Inc. \\
  \texttt{\{lizixuan,jinxiaolong,guansaiping\}@ict.ac.cn}\\
  \texttt{liwei85@baidu.com}
  }

\date{}
\begin{document}
\maketitle
\begin{abstract}
Temporal Knowledge Graphs (TKGs) have been developed and used in many different
areas. Reasoning on TKGs that predicts potential facts (events) in the future
brings great challenges to existing models. When facing a prediction
task, human beings usually search useful historical information (i.e., clues) in
their memories and then reason for future meticulously. Inspired by this
mechanism, we propose CluSTeR to predict future facts in a two-stage manner,
Clue Searching and Temporal Reasoning, accordingly. Specifically, at the clue
searching stage, CluSTeR learns a beam search policy via reinforcement learning (RL)
to induce multiple clues from historical facts. At the temporal reasoning stage,
it adopts a graph convolution network based sequence method to deduce answers
from clues. Experiments on four datasets demonstrate the substantial advantages
of CluSTeR compared with the state-of-the-art methods. Moreover, the clues found
by CluSTeR further provide interpretability for the results. 
\end{abstract}

\section{Introduction}

\begin{figure}[tbp] 
  \centering
  \includegraphics[width=3.1in]{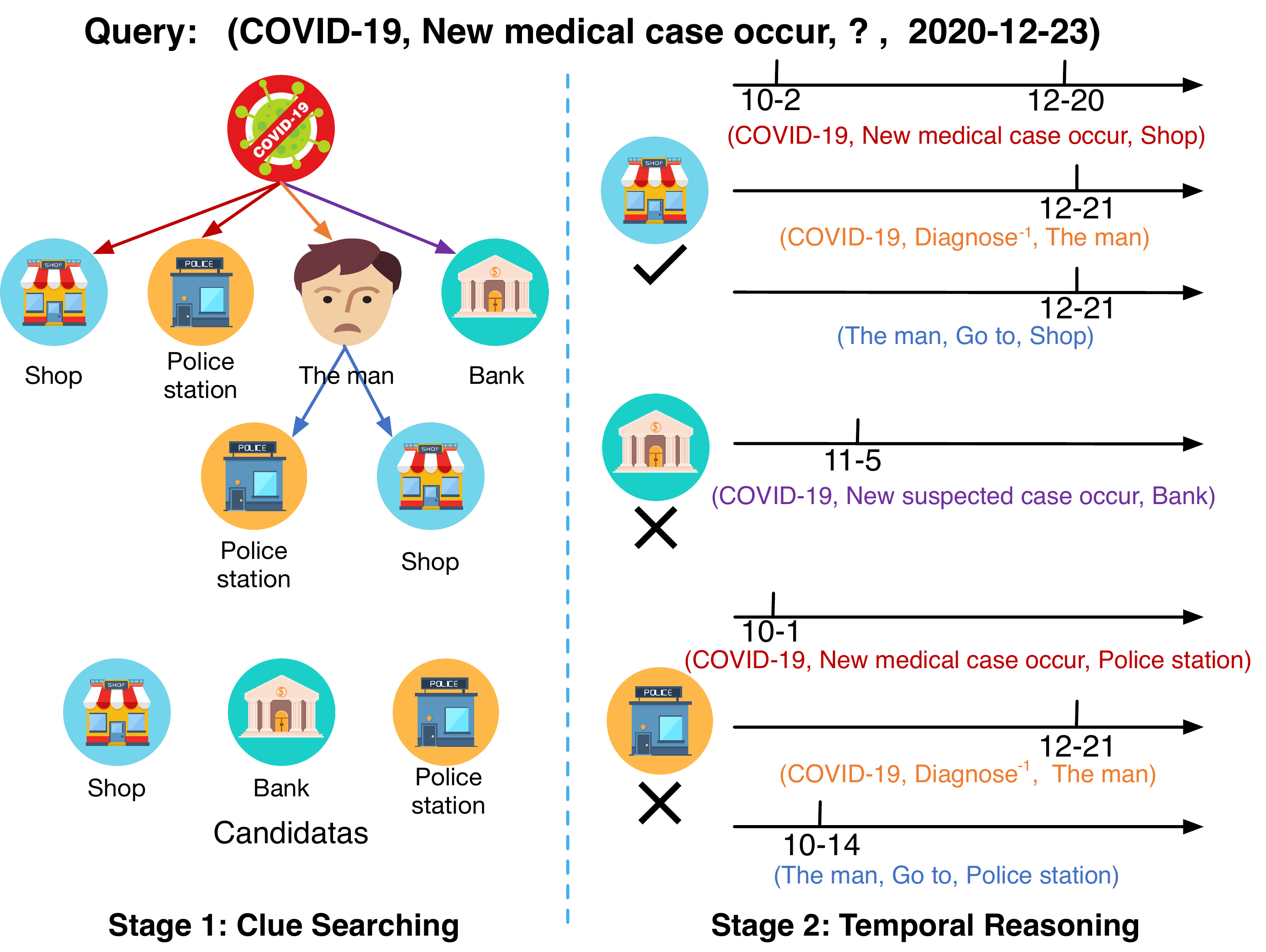}
  \caption{An illustration of the reasoning process inspired by human cognition.
  Different colors indicate different relations. $r^{-1}$ is the inverse relation of
  $r$.}
  \vspace{-5mm}
  \label{fig:case}
  \end{figure}

Temporal Knowledge Graphs
(TKGs)~\cite{boschee2015icews,gottschalk2018eventkg,gottschalk2019eventkg,
zhao2020event} have emerged as a very active research area over the last few
years. Each fact in TKGs has a timestamp indicating its time of occurrence. For
example, the fact, \emph{(COVID-19, New medical case occur, Shop, 2020-10-2)},
indicates that a new medical case of COVID-19 occurred in a shop on 2020-10-2. In
this paper, reasoning on TKGs aims to predict future facts (events) for
timestamp $t>t_{T}$, where $t_T$ is assumed to be the current
timestamp~\cite{jin2020Renet}.
An example of the task is shown in Figure~\ref{fig:case}, which attempts to
answer the query \emph{(COVID-19, New medical case occur, ?, 2020-12-23)} with
the given historical facts. Obviously, such a task may benefit many practical
applications, such as, emerging events response ~\cite{muthiah2015planned,
phillips2017using, korkmaz2015combining}, disaster
relief~\cite{signorini2011use}, and financial analysis~\cite{bollen2011twitter}.

  How do human beings predict future events? According to the dual process
  theory~\cite{evans1984heuristic, evans2003two, evans2008dual,
  sloman1996empirical}, the first thing is to search the massive-capacity
  memories and find some related historical information (i.e., clues)
  intuitively. As shown in the left part of Figure~\ref{fig:case}, there are
  mainly three categories of clues vital to the query: 1) the 1-hop paths with
  the same relation to the query (thus called repetitive 1-hop paths), such as
  \emph{(COVID-19, New medical case occur, Shop)}; 2) the 1-hop paths with
  relations different from the query (called non-repetitive 1-hop paths), such
  as \emph{(COVID-19, New suspected case occur, Bank)}; and 3) the 2-hop paths,
  such as \emph{(COVID-19, Diagnose$^{-1}$, The man, Go to, Police station)}.
  Human beings recall these clues from their memories and have some intuitive
  candidate answers for the query. Secondly, human beings get the accurate
  answer by diving deeper into the clues' temporal information and performing a
  meticulous reasoning process. As shown in the right part of
  Figure~\ref{fig:case}, the man went to the police station more than two months
  earlier than the time when he was diagnosed with \emph{COVID-19}, indicating
  that \emph{Police station} is probably not the answer. Finally, human beings
  derive the answer, \emph{Shop}. 
  
  
Existing models mainly focus on the above second process but
underestimate the first process. Some recent studies
\cite{trivedi2017know,trivedi2018dyrep} learn the evolving embeddings of
entities with all historical facts considered. However, only a few historical
facts are useful for a specific prediction. Thus, some other
studies~\cite{jin2020Renet,jin2019recurrent, zhu2020learning} mainly focus on
encoding the 1-hop repetitive paths (repetitive facts) in the history. However,
besides the 1-hop repetitive paths, there are massive other related information
in the datasets. 
Taking the widely used dataset ICEWS18~\cite{jin2020Renet} as an example, 41.2\%
of the training queries can get the answers through the 1-hop repetitive paths
in the history. But, almost 64.6\% of them can get the answers through 1-hop
repetitive and non-repetitive paths, and 86.2\% through the 1-hop and 2-hop paths.


Thus, we propose a new model called \textbf{CluSTeR}, consisting of two stages,
\textbf{Clu}e \textbf{S}earching (Stage 1) and \textbf{Te}mporal
\textbf{R}easoning (Stage 2). At Stage 1, CluSTeR formalizes clue-searching as a
Markov Decision Process (MDP)~\cite{sutton2018reinforcement} and learns a beam
search policy to solve it. At Stage 2, CluSTeR reorganizes the clues found in
Stage 1 into a series of graphs and then a Graph Convolution Network (GCN) and a
Gated Recurrent Unit (GRU) are employed to deduce accurate answers from the
graphs.


In general, this paper makes the following contributions:
 \begin{itemize}

\item We formulate the TKG reasoning task from the view of human cognition and
propose a two-stage model, CluSTeR, which is mainly composed of a RL-based clue
searching stage and a GCN-based temporal reasoning stage. 

\item We advocate the importance of clue searching for the first time,
and propose to learn a beam search policy via RL, which
can find explicit and reliable clues for the fact to be predicted.

\item Experiments demonstrate that CluSTeR achieves consistently
and significantly better performance on popular TKGs and the clues found by
CluSTeR can provide interpretability for the reasoning results.
\end{itemize}

\begin{figure*}[tbp]  
  \centering
  \includegraphics[width=6.3in]{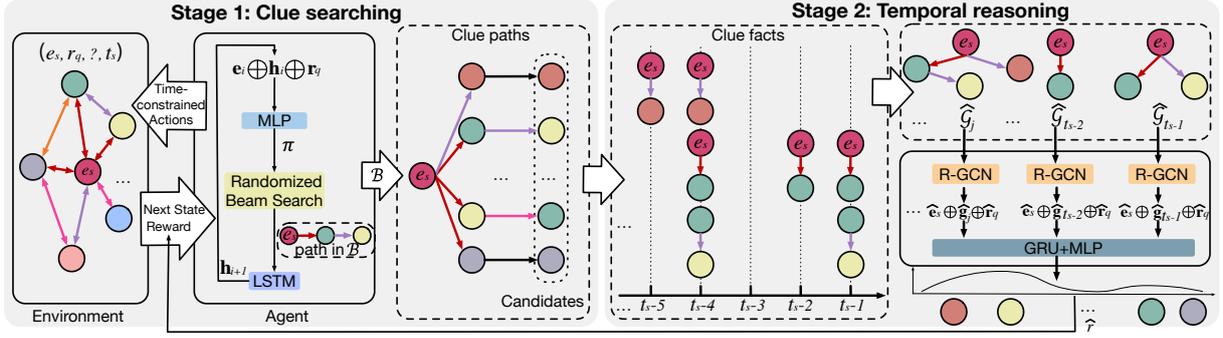}
  \caption{An illustrative diagram of the proposed CluSTeR model.}
  \label{fig:framework}
  \end{figure*}

\section{Related Work}
 
{\bf Static KG Reasoning.} Embedding based KG reasoning
models~\cite{bordes2013translating, yang2014embedding, trouillon2016complex,
dettmers2018convolutional, shang2019end, sun2018rotate} have drawn increasing
attention. All of them attend to learn the distributed embeddings for entities
and relations in KGs. Among them, some
works~\cite{schlichtkrull2018modeling,shang2019end,ye2019vectorized,vashishth2019composition}
extend GCN to relation-aware GCN for the KGs. 

However, embedding based models underestimate the symbolic compositionality of
relations in KGs, which limits their usage in more complex reasoning tasks.
Thus, some recent works~\cite{xiong2017deeppath,das2018go, lin2018multi,
chen2018variational, wang2019incorporating, li2019divine} focus on multi-hop
reasoning, which learns symbolic inference rules from relation paths. However,
all the above methods cannot deal with the temporal dependencies among facts in
TKGs.

{\bf Temporal KG Reasoning.} Reasoning on temporal KG can broadly be categorized
into two settings, interpolation~\cite{sadeghian2016temporal,
garcia2018learning, leblay2018deriving, dasgupta2018hyte,wu2019efficiently,
xu2020temporal, goel2020diachronic, wu2020temp, han2020dyernie, jung2020t} and
extrapolation~\cite{trivedi2017know, trivedi2018dyrep,han2020graph,
deng2020dynamic, jin2019recurrent,jin2020Renet, zhu2020learning,li2021temporal}, as mentioned
in~\citet{jin2020Renet}. Under the former setting, models attempt to infer
missing facts at historical timestamps.  While the latter setting, which this
paper focuses on, attempts to predict facts in the future. Orthogonal to our
work, ~\citet{trivedi2017know, trivedi2018dyrep} estimate the conditional
probability of observing a future fact via a temporal point process taking all
historical facts into consideration. Although \citet{han2020graph} extends
temporal point process to model concurrent facts, they are more capable of
modeling TKGs with continuous time, where no events may occur at the same
timestamp. Glean~\cite{deng2020dynamic} incorporates a word graph constructed by
the summary texts of events into TKG reasoning. The most related works are
RE-NET~\cite{jin2020Renet} and CyGNet~\cite{zhu2020learning}. RE-NET uses a
subgraph aggregator and GRU to model the subgraph sequence consisting of 1-hop facts. CyGNet uses a sequential copy network to model
repetitive facts. Both of them use heuristic strategies in the clue searching
stage, which may lose lots of other informative historical facts or engage some
noise. Although the above two models attempt to consider other information by
pre-trained global embeddings or an extra generation model, they still mainly
focus on modeling repetitive facts. Besides, all the models almost can not provide
interpretability for the results.

\section{The Proposed CluSTeR Model}
We start with the notations, then introduce the model as
well as its training procedure in detail.

\subsection{Notations}\label{notations}
A TKG $\mathcal{G}$ is a multi-relational directed graph with time-stamped edges
between entities. A fact in $\mathcal{G}$ can be formalized as a quadruple $(e_s, r, e_o,
t)$. It describes that a fact of relation type $r \in \mathcal{R}$ occurs
between subject entity $e_s \in \mathcal{E}$ and object entity $e_o \in
\mathcal{E}$ at timestamp $t \in \mathcal{T}$, where $\mathcal{R}$,
$\mathcal{E}$ and $\mathcal{T}$ denote the sets of relations, entities and
timestamps, respectively. TKG reasoning aims to predict the missing object
entity of $(e_{s}, r_{q}, ?, t_{s})$ or the missing subject entity of $(?,
r_{q}, e_{o}, t_{s})$ given the set of historical facts before $t_{s}$, denoted
as $\mathcal{G}_{0:t_{s}-1}$. Without loss of generality, in this paper, 
we predict the missing object entity in a fact, and the model can be easily
extended to predicting the subject entity. 

In this paper, a clue path is in the form of $(e_s, r_1, e_1, ..., r_k, e_k,
..., r_I, e_I)$, where $e_{k}\in \mathcal{E}$, $r_{k}\in \mathcal{R}$, $k=1,
..., I$, $I$ is the maximum step number and each hop in the path can be viewed
as a triple $(e_{k-1}, r_k, e_k)$. Note that, $e_0=e_s$. The clue facts are
derived from the clue paths via mapping each hop $(e_{k-1}, r_k, e_k)$ in the
paths to corresponding facts $(e_{k-1}, r_k, e_{k},t_1), (e_{k-1}, r_k,
e_{k},t_2, ...)\in \mathcal{G}_{0:t_s-1}$.

\subsection{Model Overview}
As illustrated in Figure~\ref{fig:framework}, the model consists of two stages,
clue searching and temporal reasoning. The two stages are coordinated to perform
fast and slow thinking~\cite{daniel2017thinking}, respectively, to solve the TKG
reasoning task, inspired by human cognition. Specifically, Stage 1 mainly
focuses on searching the clue paths of which the compositional semantic
information relates to the given query with the time constraints. Then, the clue
paths and the consequent candidate entities are provided for the reasoning in
Stage 2, which mainly focuses on meticulously modeling the temporal information
among clue facts and gets the final results. In the CluSTeR model, these two
stages interact with each other in the training phase and decide the final
answer jointly in the inference phase.



\subsection{Stage 1: Clue Searching}

The purpose of Stage 1 is to search and induce the clue paths related to the
given query $(e_{s}, r_{q}, ?, t_{s})$ from history. The previous
studies~\cite{jin2019recurrent,jin2020Renet,zhu2020learning} use heuristic
strategies to extract 1-hop repetitive paths, losing lots of other informative
clue paths. Besides, there are enormous facts in the history. Thus, a learnable
and efficient clue searching strategy is of great necessity. Motivated by these
observations, Stage 1 can be viewed as a sequential decision problem and
solved by the RL system. 
\subsubsection{The RL System}
The RL system consists of two parts, the agent and the environment. We formulate
the RL system as an MDP, which is a framework of learning from interactions
between the agent and the environment to find $B$ promising clue paths. Starting
from $e_s$, the agent sequentially selects outgoing edges via randomized beam
search strategy, and traverses to new entities until it reaches the maximum step
$I$. The MDP consists of the following parts:

{\bf States.} Each state $s_{i} = (e_{i}, t_{i}, e_{s}, r_{q},t_{s}) \in
\mathcal{S} $ is a tuple, where $\mathcal{S}$ is the set of all the available
states; $e_{i}$ ($e_0=e_s$) is the entity where the agent visited at step $i$; and
$t_{i}$ ($t_0=t_s$) is the timestamp of the action taken at the previous step. Note that,
$e_{s}$, $r_{q}$, and $t_{s}$ are shared by all the states for the given query.

{\bf Time-constrained Actions.} Compared to static KGs, the time dimension of
TKGs leads to an explosively large action space. Besides, the human memories
focus on the lastest occcuring events. Thus,  
we constrain the time interval between the timestamp of each fact and $t_s$ to
be no more than $m$. And the time interval between the timestamp of the previous
action and each available action is no more than $\Delta$. Therefore, the set of
the possible actions $A_i \in \mathcal{A}$ ($\mathcal{A}$ is the set of all
available actions) at step $i$ consists of the time-constrained outgoing edges
of $e_{i}$,
\begin{align}\label{eq:1}
 A_{i}=\{(r', e', t')| (e_{i}, r', e', t') &\in \nonumber\\
 \mathcal{G}_{0:t_{s}-1},\ \ |t'-t_{i}|\leq \Delta,\ \ &t_{s} - t' \leq m\}.
\end{align}

To give the agent an adaptive option
to terminate, a self-loop edge is added to $A_{i}$.  

{\bf Transition.} A transition function $\delta: \mathcal{S} \times \mathcal{A}
\rightarrow \mathcal{S}$ is deterministic under the situation of TKG and just
updates the state to new entities incident to the actions selected by the agent.

{\bf Rewards.} The agent only receives a terminal reward $R$ at the end of search,
which is the sum of two parts, binary reward and real value reward. The binary
reward is set to 1 if the destination entity $e_{I}$ is the correct target
entity $e_{o}$, and 0 otherwise. Besides, the agent gets a real value reward
$\hat{r}$ from Stage 2 if $e_I$ is the target entity, which will be introduced
in Section~\ref{temporal reasoning}.

\subsubsection{Semantic Policy Network}
Given the time-constrained action space, the compositional semantic information
implied in the clue paths and the time information of the clue facts is vital
for reasoning. 
However, considering that modeling the time information requires to dive deeply
into the complex temporal patterns of facts and is not the emphasis of Stage 1.
Thus, we design a semantic policy network which calculates the probability
distribution over all the actions according to the current state $s_i$ and
search history $h_{i} = (e_s, a_{0},..., a_{i-1})$ without considering
timestamps in Stage 1. Here, $a_{i}=(r_{i+1}, e_{i+1}, t_{i+1})$ is the action
taken at step $i = 0,...,I-1$. Note that, $h_0$ is $e_s$. Actually, the search
history without timestamps is a candidate clue path (a clue path at step $i$)
mentioned in Section~\ref{notations}.

The embedding of the action $a_i$ is $\mathbf{a}_i =\mathbf{r}_{i+1}\oplus
\mathbf{e}_{i+1}$, where $\oplus$ is the concatenation operation;
$\mathbf{r}_{i+1}, \mathbf{e}_{i+1}$ are the embeddings of $r_{i+1}$ and
$e_{i+1}$, correspondingly. Then, a Long Short Term Memory network (LSTM) is
applied to encode the candidate clue path $h_i$ as a continuous vector
$\mathbf{h}_{i}$, 
\begin{align} 
\mathbf{h}_{i} &= LSTM(\mathbf{h}_{i-1}, \mathbf{a}_{i-1}),
\end{align}
where the initial hidden embedding $\mathbf{h}_{0}$ equals to $LSTM(\mathbf{0},
\mathbf{r}_{dummy}\oplus \mathbf{e}_{s})$ and $\mathbf{r}_{dummy}$ is the embedding
of a special relation introduced to form a start action with $e_{s}$. For step
$i$, the action space is encoded by stacking the embeddings of all the actions
in $A_{i}$, which are denoted as $\mathbf{A}_{i} \in \mathbb{R}^{|A_i| \times
{2d}}$. Here, $d$ is the dimension of entity embeddings and relation embeddings.
Then, the policy network calculates the distribution $\pi$ over all the actions
by a Multi-Layer Perceptron (MLP) parameterized with $\mathbf{W}_{1}$ and
$\mathbf{W}_{2}$ as follows:
\begin{align}
\pi(a_{i}|s_{i};\! \Theta)\!=\!\eta(\!\mathbf{A}_{i}\mathbf{W}_{2}f(\mathbf{W}_{1}[\mathbf{e}_{i}\oplus\mathbf{h}_{i}\oplus\mathbf{r}_{q}]),\!\!\label{policy}
\end{align}
where $\eta(\cdot)$ is the softmax function, $f(\cdot)$ is the ReLU function~\cite{glorot2011deep} and $
\Theta$ is the set of all the learnable parameters in Stage 1.

\subsubsection{Randomized Beam Search}
In the scenario of TKGs, the occurrence of a fact may result from
multiple factors. Thus, multiple clue paths are necessary for the prediction.
Besides, the intuitive candidates from Stage 1 should recall the right answers
as many as possible. Therefore, we adopt randomized beam
search~\cite{sutskever2014sequence, guu2017language, wu2018study} as the action
sampling strategy of the agent, which injects random noise to the beam search in
order to increase the exploration ability of the agent.

Specifically, a beam contains $B$ candidate clue paths at step $i$. For each
candidate path, we append $B$ most likely actions (according to
Equation~\ref{policy}) to the end of the path, resulting in a new path pool with size
$B \times B$. Then we either pick the highest-scoring paths with probability
$\mu$ or uniformly sample a random path with probability $1-\mu$ repeatedly for
$B$ times. The score of each candidate clue path at step $i$ equals
to $\sum_{k=0}^{i}\log\pi(a_{k}|s_{k};\Theta)$. Note that, at the first step, $B$ 1-hop
candidate paths starting from $e_s$ are generated by choosing $B$ paths via
the above picking strategy.

\subsection{Stage 2: Temporal Reasoning}\label{temporal reasoning} To dive
deeper into the temporal information among clue facts at different timestamps
and the structural information among concurrent clue facts, Stage 2 reorganizes
all clue facts into a sequence of graphs $\hat{\mathcal{G}} =
\{\hat{\mathcal{G}}_0, ..., \hat{\mathcal{G}}_{j},
...,\hat{\mathcal{G}}_{t_{s}-1}\}$, where each $\hat{\mathcal{G}}_{j}$ is a
multi-relational graph consisting of clue facts at timestamp $j=0,...t_s-1$. We use an
$\omega$-layer RGCN~\cite{schlichtkrull2018modeling} to model
$\hat{\mathcal{G}}_{j}$,


%
\begin{equation} \label{eq:rgcn}
  \hat{\mathbf{h}}_{o, j}^{l+1}=f\!\!\left(\!\!\frac{1}{d_{o}}\!\sum_{(s, r)|(s, r, o, j)\in \hat{\mathcal{G}}_{j} }\!\!\mathbf{W}_{r}^{l}\hat{\mathbf{h}}_{s, j}^{l}
   \!+\!  \mathbf{W}_{loop}^{l}\hat{\mathbf{h}}_{o,j}^{l}\!\!\right)\!, 
  \end{equation}
where $\hat{\mathbf{h}}_{o,j}^{l}$ and $\hat{\mathbf{h}}_{s,j}^{l}$ denote the $l^{th}$
layer embeddings of entities $o$ and $s$ in $\hat{\mathcal{G}}_{j}$ at timestamp
$j$, respectively; $\mathbf{W}_{r}^{l}$ and $\mathbf{W}_{loop}^{l}$ are the weight
matrices for aggregating features from different relations and self-loop in the
$l^{th}$ layer; $d_{o}$ is the in-degree of entity $o$; the input embedding
for each entity $k$, $\hat{\mathbf{h}}_{k,j}^{l=0}$ is set to $\hat{\mathbf{e}}_k$
, which is different from that of Stage 1. 

Then, $\hat{\mathbf{g}}_{j}$, the embedding of $\hat{\mathcal{G}}_{j}$, is
calculated by the mean pooling operation of all entity embeddings calculated by
Equation~\ref{eq:rgcn} in $\hat{\mathcal{G}}_{j}$. The concatenation of
$\hat{\mathbf{e}}_{s}$, $\hat{\mathbf{g}}_{j}$ and $\hat{\mathbf{r}}_{q}$ (the
embedding of $r_q$ in Stage 2) is fed into a GRU,
\begin{equation}\label{eq:rnn}
\mathbf{H}_{j} = GRU([\hat{\mathbf{e}}_{s}\oplus\hat{\mathbf{g}}_{j}\oplus\hat{\mathbf{r}}_{q}], \mathbf{H}_{j-1}).
\end{equation}

The final output of GRU, denoted as $\mathbf{H}_{t_{s}-1}$, is fed into a MLP
decoder parameterized with $\mathbf{W}_{mlp}$ to get the final scores for all the
entities, i.e.,
\begin{equation}\label{eq:final_p}
p(e|e_{s},r_{q},t_{s})
= \sigma (\mathbf{H}_{t_{s}-1}^{T} \cdot \mathbf{W}_{mlp}),
\end{equation}
where $\sigma$ is the sigmoid activation function.

Finally, we re-rank the candidate entities according to
Equation~\ref{eq:final_p}. To give a positive feedback to the clue paths
arriving at the answer, Stage 2 gives a beam-level reward which
equals to the final score of $e_{I}$ from Equation~\ref{eq:final_p}, i.e, $\hat{r} = p(e_{I})$, to Stage 1.


\subsection{Training Strategy}
For Stage 1, the beam search policy network is trained by maximizing the
expected reward over all queries in the training set,
\begin{equation}
\mathcal{J}(\Theta) \!\!= \!\!\mathbb{E}_{(e_s, r_q, e_o, t_s)\in \mathcal{G}}[\mathbb{E}_{a_0,...a_{I-1}}[R(e_I|e_s,r_q,t_s)]]. \label{object1}
\end{equation}

The REINFORCE algorithm~\cite{williams1992simple} is used to optimize
Equation~\ref{object1}. For Stage 2, we define the objective function using
cross-entropy:
\begin{equation}
  \mathcal{J}(\Phi) \!=\!-\frac{1}{|\mathcal{G}|}\sum_{(e_s, r_q, e_o, t_s)\in \mathcal{G}}\!\!\log p(e_o|e_s,r_q,t_s), \label{object2}
\end{equation}
where $\Phi$ is the set of all the learnable parameters in Stage 2. The
Adam~\cite{kingma2014adam} optimizer is used to minimize Equation~\ref{object2}.
As Stages 1 and Stage 2 are correlated mutually, they are trained jointly. Stage
1 is pre-trained with only binary reward before the joint training process
starts. Then Stage 2 is trained with the parameters of Stage 1 frozen. At last,
we jointly train the two stages. Such a training strategy is widely used by
other RL studies~\cite{bahdanau2016actor,feng2018reinforcement}.

\section{Experiment}
We design experiments to answer the following questions: \textbf{Q1}. How does
CluSTeR perform on the TKG reasoning task? \textbf{Q2}. How do the two
stages contribute to the final results respectively? \textbf{Q3}. Which clues are
found and used for reasoning? \textbf{Q4}. Can CluSTeR provide some
interpretability for the results? 

\subsection{Experimental Setup} \label{metrics}
{\bf Datasets and Metrics.} There are four typical TKGs commonly used in
previous studies, namely, ICEWS14~\cite{garcia2018learning},
ICEWS05-15~\cite{garcia2018learning}, ICEWS18~\cite{jin2019recurrent} and
GDELT~\cite{jin2020Renet}. The first three datasets are from the Integrated
Crisis Early Warning System (ICEWS)~\cite{boschee2015icews} and the last one is
from Global Database of Events, Language, and Tone
(GDELT)~\cite{leetaru2013gdelt}. We evaluate CluSTeR on all these datasets.
ICEWS14 and ICEWS05-15 are divided into training, validation, and test sets
following the preprocessing on ICEWS18 in RE-NET~\cite{jin2020Renet}. The
details of the datasets are presented in Table~\ref{table:datasets}.

\begin{table}
\small
\centering
\begin{tabular}{lrrrrrr}
\toprule
Datasets           &ICE14   &ICE05-15  &ICE18  &GDELT\\
\midrule
$\#\mathcal{E}$              &6,869    &10,094  &23,033   &7,691\\
$\#\mathcal{R}$               &230     &251   &256     &240 \\
$\#{Train}$     &74,845   &368,868  &373,018  &1,734,399\\
$\#{Valid}$      &8,514    &46,302  &45,995   &238,765\\
$\#{Test}$      &7,371    &46,159  &49,545    &305,241\\
Time gap         &1 day   &1 day   &1 day  &15 mins\\

\bottomrule
\end{tabular}
\vspace{-2mm}
\caption{Statistics of the datasets.}
\vspace{-6mm}
\label{table:datasets}
\end{table}

In the experiments, the widely used Mean Reciprocal Rank (MRR) and Hits@\{1,10\}
are employed as the metrics. Without loss of generality, only the experimental
results under the raw setting are reported. The filtered setting is
not suitable for the reasoning task under the exploration setting, as mentioned
in~\cite{han2020graph, ding2021temporal, jain2020temporal}. 
The reason is explained in
terms of an example as follows: Given a test quadruple (Barack Obama, visit,?,  2015-1-25) with the correct answer India. Assume there is a quadruple (Barack
Obama, visit, Germany, 2013-1-18) in the training set. The filtered setting
used in the previous studies ignores time information and considers (Barack
Obama, visit, Germany, 2015-1-25) to be valid because (Barack Obama, visit,
Germany, 2013-1-18) appears in the training set. It thus removes the quadruple
from the corrupted ones. However, the fact (Barack Obama, visit, Germany) is
temporally valid on 2013-1-18, instead of 2015-1-25. Therefore, to test
the quadruple (Barack Obama, visit,?, 2015-1-25), (Barack Obama, visit,
Germany, 2015-1-18) should not be removed. In this way, the filtered setting
wrongly removes quite a lot of quadruples and thus leads to over-optimistic
experimental performance.

\begin{table*}[htb]
\small
\centering
\begin{tabular}{lcccccccccccc}
    
\toprule
\multirow{2}{*}{Model} &\multicolumn{3}{c}{ICE14} &\multicolumn{3}{c}{ICEWS05-15} &\multicolumn{3}{c}{ICE18} &\multicolumn{3}{c}{GDELT}\\
\cmidrule(r){2-4}  \cmidrule(r){5-7} \cmidrule(r){8-10} \cmidrule(r){11-13}  &MRR &H@1 &H@10
&MRR &H@1 &H@10 &MRR &H@1 &H@10 &MRR &H@1 &H@10\\
\midrule
\scriptsize{DistMult}\!\!\!  &24.9   &17.3  &40.2  &16.4 &9.8  &29.9  & 17.5  & 10.1 &32.6 &15.6 &9.3   &28.0        \\
\scriptsize{ComplEx}\!\!\!   &31.9   &22.2  &50.7  &23.1 &14.5 &40.6  & 18.8  & 11.1 &26.8 &12.3 &8.0   &20.6    \\
\scriptsize{RGCN}\!\!\!     &27.1   &18.4  &44.2  &27.3 &19.1 &43.6  & 17.0  & 8.7  &34.0 &10.9 &4.6   &22.6        \\
\scriptsize{ConvE}\!\!\!     &30.9   &21.7  &50.1  &25.2 &16.0 &44.4  & 24.8  &15.1 &44.9 &17.3 &10.4  &31.3   \\
\scriptsize{RotatE}\!\!\!    &27.5   &18.0  &47.2  &19.9 &10.9 &38.7  & 15.5  & 7.0  &33.9 &5.3 &1.2  &12.5   \\
\scriptsize{MINERVA}\!\!\!    &33.2   &25.7  &48.3  &30.7 &25.8 &39.9  & 21.0  & 15.3  &33.0 &12.1 &10.0  &16.7   \\

\midrule
\scriptsize{Know-Evolve}\!\!\!  &-- &-- &--    &-- &-- &-- &7.4 &3.3 &14.8 &15.9 &11.7  &22.3 \\
\scriptsize{DyRep}\!\!\!     &-- &-- &--  &-- &-- &-- &7.8 &3.6 &16.3 &16.3 &\textbf{11.8} &23.9\\
\scriptsize{RGCRN}\!\!\!     &36.9    &27.0    &56.1    &39.4    &28.7    &60.4    &26.2    &16.4    &45.8    &17.7    &10.9    &30.9\\
\scriptsize{EvolveRGCN}\!\!\!    &37.1    &27.0    &57.0    &40.7  &30.3
&61.3    &23.6    &36.3    &50.4    &17.4 &11.0 &29.9\\
\scriptsize{CyGNet}\!\!\!  &36.5  &27.4  & 54.4  &37.4  &27.5  &56.1 &26.8  &17.1  &45.7   &18.0  &10.9   &31.6\\
\scriptsize{RE-NET}\!\!\!  &38.9  &29.3  &57.5   &41.7  &31.1  &62.0 &28.4  &18.4  &47.9   &\textbf{19.0}  & 11.6  &\textbf{33.5}\\
\scriptsize{CluSTeR}\!\!\! &\textbf{46.0}  &\textbf{33.8}  &\textbf{71.2}   &
\textbf{44.6}  & \textbf{34.9}  & \textbf{63.0}  & \textbf{32.3}  &
\textbf{20.6}   & \textbf{55.9} & 18.3  & 11.6   & 31.9\\
\bottomrule
\end{tabular}
\vspace{-2mm}
\caption{Experimental results on TKG reasoning (in percentage) compared with static models (the top part) and temporal models (the bottom part).}
\vspace{-2mm}
\label{table:results}
\end{table*}

\begin{table}
\small
\centering
\setlength{\tabcolsep}{0.3em}
\begin{tabular}{lcccc}
\toprule
Model   & ICE14  & ICE05-15  & ICE18  & GDELT \\
\midrule
Stage 1 ($I=2$)       &43.1  &43.3 &27.6 &15.3\\
Stage 1 ($I=1$)       &44.1  &46.0 &30.3 &17.6\\
Stage 2        &41.5  &45.0 &30.1 &\textbf{19.6}    \\
CluSTeR               &\textbf{46.8}  &\textbf{46.9} &\textbf{33.1} &18.7\\
\bottomrule
\vspace{-4mm}
\end{tabular}
\caption{Results (in percentage) by different variants of CluSTeR on all the datasets.}
\label{table:ablation}
\vspace{-4mm}
\end{table}
  
{\bf Baselines.} The CluSTeR model is compared with two categories of models,
i.e., models for static KG reasoning and models for TKG reasoning under the
exploration setting. The typical static models
DistMult~\cite{yang2014embedding}, ComplEx~\cite{trouillon2016complex},
RGCN~\cite{schlichtkrull2018modeling}, ConvE~\cite{dettmers2018convolutional}
and RotaE~\cite{sun2018rotate} are selected with the temporal information of
facts ignored. We also choose MINERVA~\cite{das2018go}, the RL-based multi-hop
reasoning model, as the baseline. For TKG models, the representative
Know-evolve~\cite{trivedi2017know}, DyRep~\cite{trivedi2018dyrep},
CyGNet~\cite{zhu2020learning} and RE-NET~\cite{jin2020Renet} are selected.
Besides, following RE-NET~\cite{jin2020Renet}, we extend two models for temporal
homogeneous graphs, GCRN~\cite{seo2018structured} and
EvolveGCN-O~\cite{pareja2019evolvegcn}), to RGCRN and EvolveRGCN by replacing
GCN with RGCN. We use ConvE~\cite{dettmers2018convolutional}, a more stronger
decoder to replace the MLP~\cite{jin2020Renet} for the two models. For
Know-evolve and DyRep, RE-NET extends them to TKG reasoning task but does not
release their codes. Thus, we only report the results from their papers. For
other baselines, we reproduce all the results with the optimal parameters tuning
on the validation set.

{\bf Implementation Details.}  In the experiments, the embedding dimension $d$
for the two stages, is set to 200. For Stage 1, we adopt an adaptive approach for
selecting the time interval $m$. Specifically, for ICEWS14, ICEWS05-15, and
GDELT, $m$ is set to the last one timestamp the query pattern ($e_s$, $r_q$, ?)
appearing in the dataset before $t_s$. And for ICEWS18, $m$ is set to the last
third timestamp. $\Delta$ is set to 3 for all the datasets. We set the maximum
step number $I={1,2}$ and find $I=1$ is better for all the datasets. The number
of the LSTM layers is set to 2 and the dimension of the hidden layer of LSTM is
set to 200 for all the datasets. The beam size is set to 32 for the three ICEWS
datasets and 64 for GDELT. $\mu$ is set to 0.3 for all the datasets. For Stage
2, the maximum sequence length of GRU is set to 10, the number of
the GRU layers is set to 1 and the number of the RGCN layers is set to 2 for all the 
datasets. For each fact in $\mathcal{G}_{0:t_s-1}$, we add the corresponding
inverse fact into $\mathcal{G}_{0:t_s-1}$. All the experiments are carried out on Tesla V100.

\subsection{Results on TKG Reasoning}\label{Experimental Results} The results on
TKG reasoning are presented in Table~\ref{table:results}. CluSTeR consistently
outperforms the baselines on all the ICEWS datasets, which convincingly verifies
its effectiveness and answers Q1. Especially on ICEWS14, CluSTeR even
achieves the improvements of 7.1\% in MRR, 4.5\% in Hits@1, and 13.7\% in
Hits@10 over the best baselines. Specifically, CluSTeR significantly outperforms
the static models (i.e., those in the first block of Table~\ref{table:results})
because it captures the temporal information of some important history.
Moreover, CluSTeR drastically performs better than those temporal models.
Compared with DyRep and Know-evolve that consider all the history, CluSTeR can
focus on more vital clues. Different from RGCRN and EvolveRGCN which model all
history from several latest timestamps, CluSTeR models a longer history after
reducing all history to a few clues.
CyGNet and RE-NET mainly focus on modeling the repetitive clues or all the
1-hop clues and show strong performance. CluSTeR also outperforms them on the three ICEWS
datasets, because the RL-based Stage 1 can find more explicit and reliable
clues. 

\begin{table*}
  \small
  \centering
  \begin{tabular}{cccc}
  \toprule
  &Query relations & 2-hop paths  &Scores \\
  \midrule
    & (A, Declare ceasefire, C) &(A, Intent to cooperate, B, Intent to meet, C)
    &0.4071\\
    & (A, Intent to settle dispute,C) & (A, Consult, B, Intent to diplomatic
    cooperation, C) &0.3843 \\
    & (A, Intent to settle dispute, C) &(A, Intent to diplomatic cooperation, B,
  Intent to meet, C) &0.3725 \\
    &(A, Halt negotiations, C) & (A, Engage in negotiation, B, Intent to meet,C)
    &0.3717 \\
    & (A, Accuse of crime, C) &(A, Accuse, B, Criticize or denounce, C)  &0.3256
    \\
  \bottomrule
  \end{tabular}
  \vspace{-2mm}
  \caption{The top five convincing 2-hop paths extracted by AMIE+ from ICEWS18.}
  \label{table:clue_AMIE}
  \vspace{-2mm}
  \end{table*}

The experimental results on GDELT demonstrate that the performance of static
models and temporal ones are similarly poor, as compared with those of the other
three datasets. We further analyze the GDELT dataset and find that a large
number of its entities are abstract concepts which do not indicate a specific
entity (e.g., PRESIDENT, POLICE and GOVERNMENT). Among the top 50 frequent
entities, 28 are abstract concepts and 43.72\% corresponding events involve
abstract concepts. Those abstract concepts make future prediction under the raw
setting almost impossible, since we cannot predict a president's activities
without knowing which country he belongs to. 

\subsection{Ablation Study}
To answer Q2, i.e., how the two stages contribute to the final results,
we report the MRR results of the variants of CluSTeR on the validation set of
all the datasets in Table~\ref{table:ablation}. The first two lines of
Table~\ref{table:ablation} show the results only using Stage 1, where the
maximum step $I$ is set to 1 and 2, respectively.
Following~\citet{lin2018multi}, the score of the target entity is set to the
highest score among the paths when more than one path leads to it.  It can be
observed that the results decrease when only using Stage 1, because the temporal
information among facts is ignored. The third line shows the results only using
Stage 2 with extracted 1-hop repetitive clues as the inputs. The results decrease on
all the ICEWS datasets when only using Stage 2, demonstrating that only
repetitive clues are not enough for the prediction. For GDELT, only Stage 2
achieves the best results, which demonstrates that only using repetitive clues
is effective enough for it. It is because that only using the most
straightforward repetitive clues in Stage 2 can alleviate the influence of noise
produced by abstract concepts. It also matches our
observations mentioned in Section~\ref{Experimental Results}.

From the first two lines of Table~\ref{table:ablation}, it can be seen
that the performance of Stage 1 decreases when $I$ is set to 2. To further
analyze the reason, we extract paths from ICEWS18 without considering timestamps
via AMIE+~\cite{galarraga2015fast}, a widely used and accurate approach to
extract logic rules (paths) in static KGs. We check the top fifty paths manually and present the top five convincing paths in
Table~\ref{table:clue_AMIE}. It can be seen that there are no strong
dependencies between the query relations and the 2-hop paths. Thus, in this
situation, longer paths bring exponential noise clues, which pull down the
precision. We do experiments on all the datasets from ICEWS and GDELT and find
the same conclusion. We leave it for future work to construct a more complex
dataset for verifying the effectiveness of multi-hop clue paths.

\begin{figure}[tbp]
  \centering
  \includegraphics[width=3in]{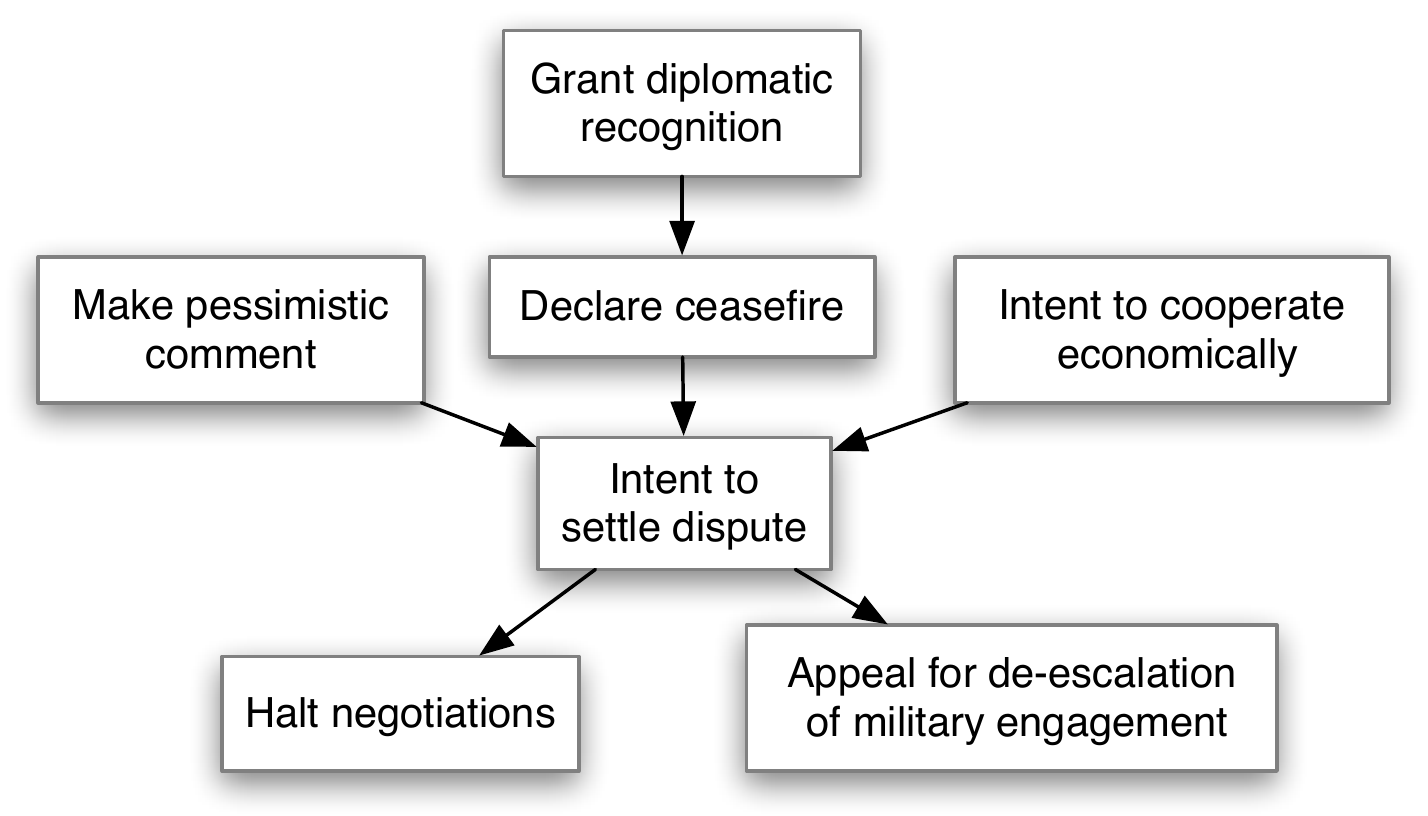}
  \caption{A clue graph constructed by Stage 1. }
  \label{fig:clue_graph}
  \end{figure}

\begin{figure}[tbp] 
  \centering
  \includegraphics[width=3in]{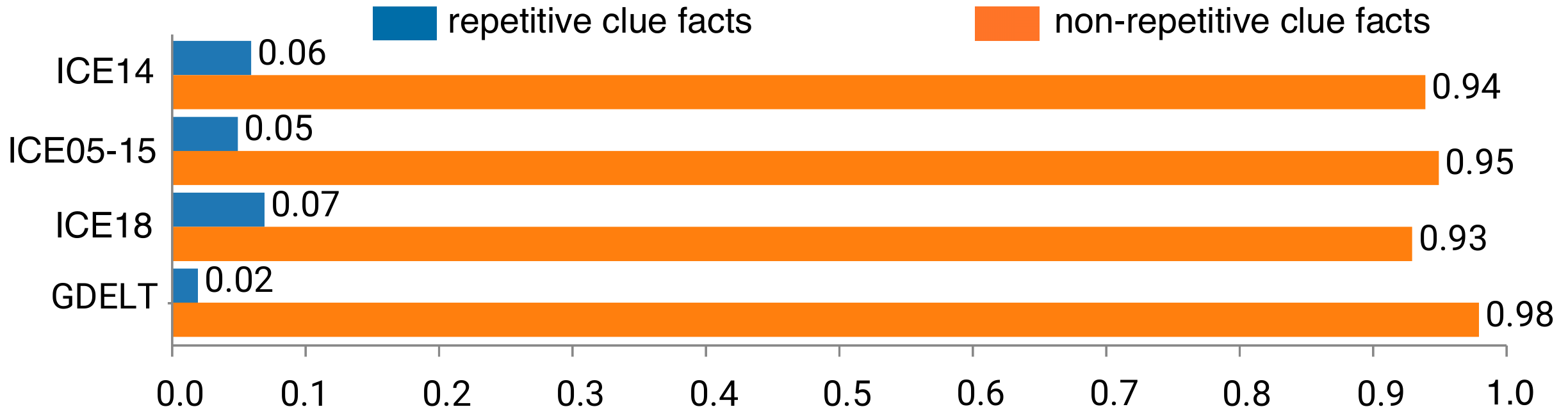}
  \caption{Statistic of categories of clue facts in Stage 2.}
  \label{fig:clue_kind}
  \vspace{-4mm}
  \end{figure}

\begin{figure*}[tbp]
\centering
\includegraphics[width=6in]{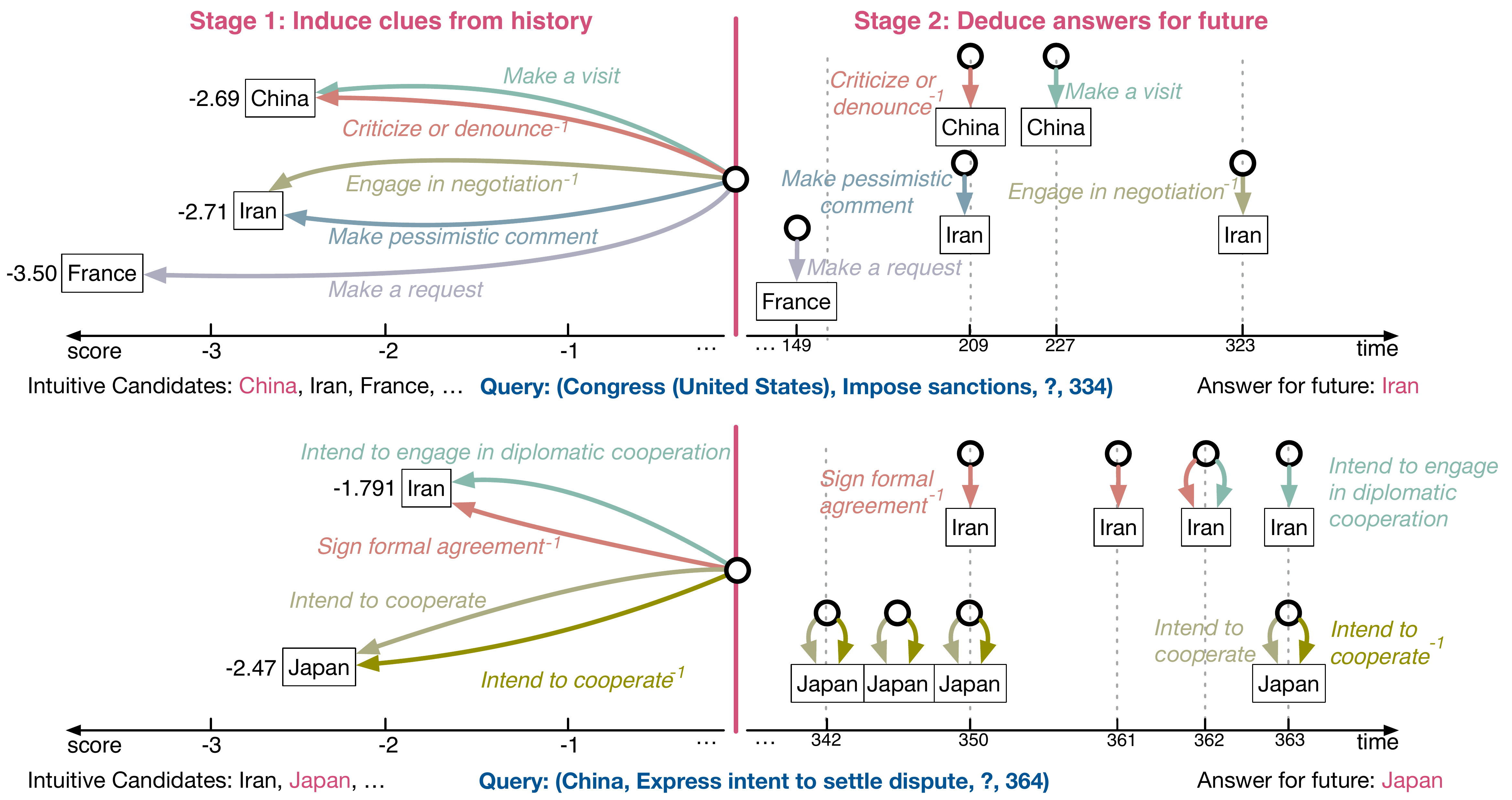}
\vspace{-2mm}
\caption{Two cases to illustrate how CluSTeR conducts reasoning and explains the
results. Each black circle represents a query entity.}
\label{fig:2-stage-case}
\end{figure*}

\subsection{Detail Analysis}\label{Detail Analysis} To answer Q3, we
show some non-repetitive clues found in Stage 1 in
Figure~\ref{fig:clue_graph}. We use (relation in 1-hop
non-repetitive clue path, query relation) pairs on ICEWS18 to
construct a clue graph. Arrows begin with the relations in the clue paths
and end with the query relations. It is interesting to find that CluSTeR can actually
find some causal relations. Moreover, compared to the 2-hop clue paths shown in
Table~\ref{table:clue_AMIE}, the 1-hop clue paths are more informative. It also
gives explanations to the outperformance of the 1-hop clue paths. 

Besides, we illustrate the statistics of clue facts used during Stage 2 in
Figure~\ref{fig:clue_kind}. The proportion of the repetitive clue facts is less
than 7\% and the proportion of the non-repetitive clue facts is more than 93\%
on the datasets. The abundant of the non-repetitive clue facts used in Stage 2 also
explains the outperformance of CluSTeR to a certain degree.




\subsection{Case Study}
To answer Q4, we show how CluSTeR conducts reasoning and explains the
results for the given two queris from the test set of ICEWS14 in
Figure~\ref{fig:2-stage-case}. 
For the first query: \emph{(Congress (United
States), Impose sanctions, ?, 334\footnote{Here, 334 represents the 334th day in
the year 2014.})}, we choose the top three candidates in Stage 1 and demonstrate
some clue paths of the three entities in the left top part of
Figure~\ref{fig:2-stage-case}. 
The clue paths like \emph{(Congress (United
States), Criticize or denounce$^{-1}$, China)}, \emph{(Congress (United States),
Engage in negotiation$^{-1}$, Iran)} give the evidence for candidate entities
\emph{China} and \emph{Iran}, correspondingly. 
In Stage 1, CluSTeR has an intuitive candidate
set including \emph{China}, \emph{Iran} and \emph{France}. 
The score of \emph{China} (-2.69) and \emph{Iran}
(-2.71) are similar but the wrong answer, \emph{China}, has a higher score than the
right one, \emph{Iran}. 
It is because Stage 1 does not take the temporal information
into consideration. 
However, the score gap is obvious between \emph{Iran} and \emph{France},
which shows that Stage 1 can measure the qualities of different clue paths and
distinguish the semantic-related entities from the others. In Stage 2, CluSTeR
reorganizes the clue facts by their timestamps, as shown in the right top part
of Figure~\ref{fig:2-stage-case}. \emph{(Congress (United State), Engage in
negotiation$^{-1}$, Iran, 323)} and
\emph{(Congress (United State), Make a visit, China, 227)}
make \emph{Iran} the more possible answer. For the second query: \emph{(China, Express intent to settle dispute, ?, 364)},
clue paths in the left bottom of Figure~\ref{fig:2-stage-case} are all
associated with the query. Stage 1 induces all entities to only two entities
through these clue paths but misleads to the wrong answer, \emph{Iran}.
Actually, even a human may give the wrong answer with only fasting thinking.
After diving into the temporal information of clue facts and conduct slow
thinking, some causal information and period information can be captured by Stage 2. Although \emph{Sign formal agreement} is associated with \emph{Express intent to
settle dispute}, it can not be the reason for the latter. Moreover, from the
subgraph sequence in the right bottom part of Figure~\ref{fig:2-stage-case}, it
can be seen that the cooperation period between \emph{China} and \emph{Japen}
just begins at 363, but the cooperation period between \emph{China} and
\emph{Iran} has been going on for several days. \emph{(China, Express
intent to settle dispute, ?, 364)} is more likely to be an antecedent event to
the cooperation period and the answer is \emph{Japen}. 

Above all, for each fact to be predicted, CluSTeR can provide the clues for each
candidate entity, which presents the insight and provides interpretability for
the reasoning results. It is similar to the natural thinking pattern of human,
in which only explicit and reliable clues are needed.

\begin{table*}[htb]
  \small
  \centering
  \begin{tabular}{lcccccccccccc}
      
  \toprule
  \multirow{2}{*}{Model} &\multicolumn{3}{c}{ICE14} &\multicolumn{3}{c}{ICEWS05-15} &\multicolumn{3}{c}{ICE18} &\multicolumn{3}{c}{GDELT}\\
  \cmidrule(r){2-4}  \cmidrule(r){5-7} \cmidrule(r){8-10} \cmidrule(r){11-13}  &MRR &H@1 &H@10
  &MRR &H@1 &H@10 &MRR &H@1 &H@10 &MRR &H@1 &H@10\\
    \midrule
  raw\!\!\! &46.0  &{33.8}  &{71.2}   &{44.6}  & {34.9}  & {63.0}  & {32.3}  &{20.6}   & {55.9} & 18.3  & 11.6   & 31.9\\
  filtered\!\!\! &{47.1}  &{35.0}  &{72.0}   &{45.4}  & {34.3}  & {67.7}  & {34.5}  &{22.9}   & {57.7} & 18.5  & 12.1   & 32.1\\
  \bottomrule
  \end{tabular}
  \vspace{-2mm}
  \caption{Experimental results under the raw setting and the (time-aware) filter setting.}
  \label{table:time-filtered results}
  \end{table*}

\subsection{Performance under the Time-aware Filtered Setting}
As mentioned in Section~\ref{metrics}, the widely adopted filtered setting in
the existing studies is not suitable for the temporal reasoning task
addressed in this paper. The essential problem of the above filtered setting is
that it ignores the time information of a fact. Therefore, we also adopt an
improved filtered setting where the time information is also considered, thus
called time-aware filtered setting~\cite{han2020graph, hanexplainable}. Specifically, only the
facts occur at the predicted time are filtered. The results are in
Table~\ref{table:time-filtered results}. It can been seen that the experimental
results under the time-aware filtered setting are close to those under the raw
setting. This is because that only a very small number of facts are removed
under this filtered setting. The results also show the convincing of the
raw setting.

\section{Conclusions}
In this paper, we proposed a two-stage model from the view of human cognition,
named CluSTeR, for TKG reasoning. CluSTeR consists of a RL-based clue searching
stage (Stage 1) and a GCN-based temporal reasoning stage (Stage 2). In Stage 1,
CluSTeR finds reliable clue paths from history and generates intuitive candidate
entities via RL.  With the found clue paths as input, Stage 2 reorganizes the
clue facts derived from the clue paths into a sequence of graphs and performs
deduction on them to get the answers. By the two stages, the model demonstrates
substantial advantages on TKG reasoning. Finally, it should be mentioned that, although the four TKGs adopted in the experiments were created based on the events in the real world, the motivation of this paper is to propose this TKG reasoning model only for scientific research.

\section*{Acknowledgment}
We gratefully acknowledge the help and assistance from Long Bai, Yunqi Qiu, Bing
Li and Bingbing Xu. Moreover, the work is supported by the National Key Research and
Development Program of China under grant 2016YFB1000902, the National Natural
Science Foundation of China under grants U1911401, 62002341, 61772501, U1836206 and 61722211, the GFKJ Innovation Program, Beijing Academy of
Artificial Intelligence under grant BAAI2019ZD0306, and the Lenovo-CAS Joint Lab
Youth Scientist Project.

\bibliographystyle{acl_natbib}
\bibliography{acl2021}


\end{document}